\documentclass[10pt,twocolumn,letterpaper]{article}

\usepackage{iccv}              
\usepackage{graphicx}
\usepackage{amsmath}
\usepackage{amssymb}
\usepackage{booktabs}
\usepackage{pifont}
\usepackage{color}

\usepackage{comment} 
\definecolor{iccvblue}{rgb}{0.21,0.49,0.74}
\usepackage[pagebackref,breaklinks,colorlinks,allcolors=iccvblue]{hyperref}

\def\paperID{3673} 
\def\confName{ICCV}
\def\confYear{2025}
\usepackage{multirow}
\title{AU-Blendshape for Fine-grained Stylized 3D Facial Expression Manipulation}

\author{Hao Li$^{1,2}$, Ju Dai$^{2,}$\thanks{Corresponding authors}, Feng Zhou$^{3}$, Kaida Ning$^{2}$, Lei Li$^{4,5}$, Junjun Pan$^{1,2,*}$\\
$^{1}$Beihang University,  $^{2}$Peng Cheng Laboratory, $^{3}$North China University of Technology, \\$^{4}$University of Washington, $^{5}$University of Copenhagen\\
{\tt\small \{lih09, daij, ningkd\}@pcl.ac.cn, zhoufeng@ncut.edu.cn, lilei@di.ku.dk, pan\_junjun@buaa.edu.cn} 
}

\begin{document}
\maketitle
\begin{abstract}

While 3D facial animation has made impressive progress, challenges still exist in realizing fine-grained stylized 3D facial expression manipulation due to the lack of appropriate datasets. In this paper, we introduce the AUBlendSet, a 3D facial dataset based on AU-Blendshape representation for fine-grained facial expression manipulation across identities. AUBlendSet is a blendshape data collection based on 32 standard facial action units (AUs) across 500 identities, along with an additional set of facial postures annotated with detailed AUs. Based on AUBlendSet, we propose AUBlendNet to learn AU-Blendshape basis vectors for different character styles. AUBlendNet predicts, in parallel, the AU-Blendshape basis vectors of the corresponding style for a given identity mesh, thereby achieving stylized 3D emotional facial manipulation. We comprehensively validate the effectiveness of AUBlendSet and AUBlendNet through tasks such as stylized facial expression manipulation, speech-driven emotional facial animation, and emotion recognition data augmentation. Through a series of qualitative and quantitative experiments, we demonstrate the potential and importance of AUBlendSet and AUBlendNet in 3D facial animation tasks. To the best of our knowledge, AUBlendSet is the first dataset, and AUBlendNet is the first network for continuous 3D facial expression manipulation for any identity through facial AUs.  Our source code is available at \href{https://github.com/wslh852/AUBlendNet.git}{https://github.com/wslh852/AUBlendNet.git}.
\end{abstract}    
\section{Introduction}
\label{sec:intro}

Along with the development of virtual reality and metaverse, 3D facial animation has become a prominent research topic~\cite{CudeiroBLRB19,FanLSWK22,FaceDiffuser}. Facial expressions, as the core expression of human emotions and intentions, play a crucial role in facial animation production. A fine-grained, controllable facial expression manipulation algorithm with customized stylization can effectively help animators accurately construct characters' emotions, thereby improving narrative effectiveness and audience resonance~\cite{WangZYYJ20,WangDYSW23}.

\begin{figure}[htb]
 \centering 
 \includegraphics[width=\columnwidth]{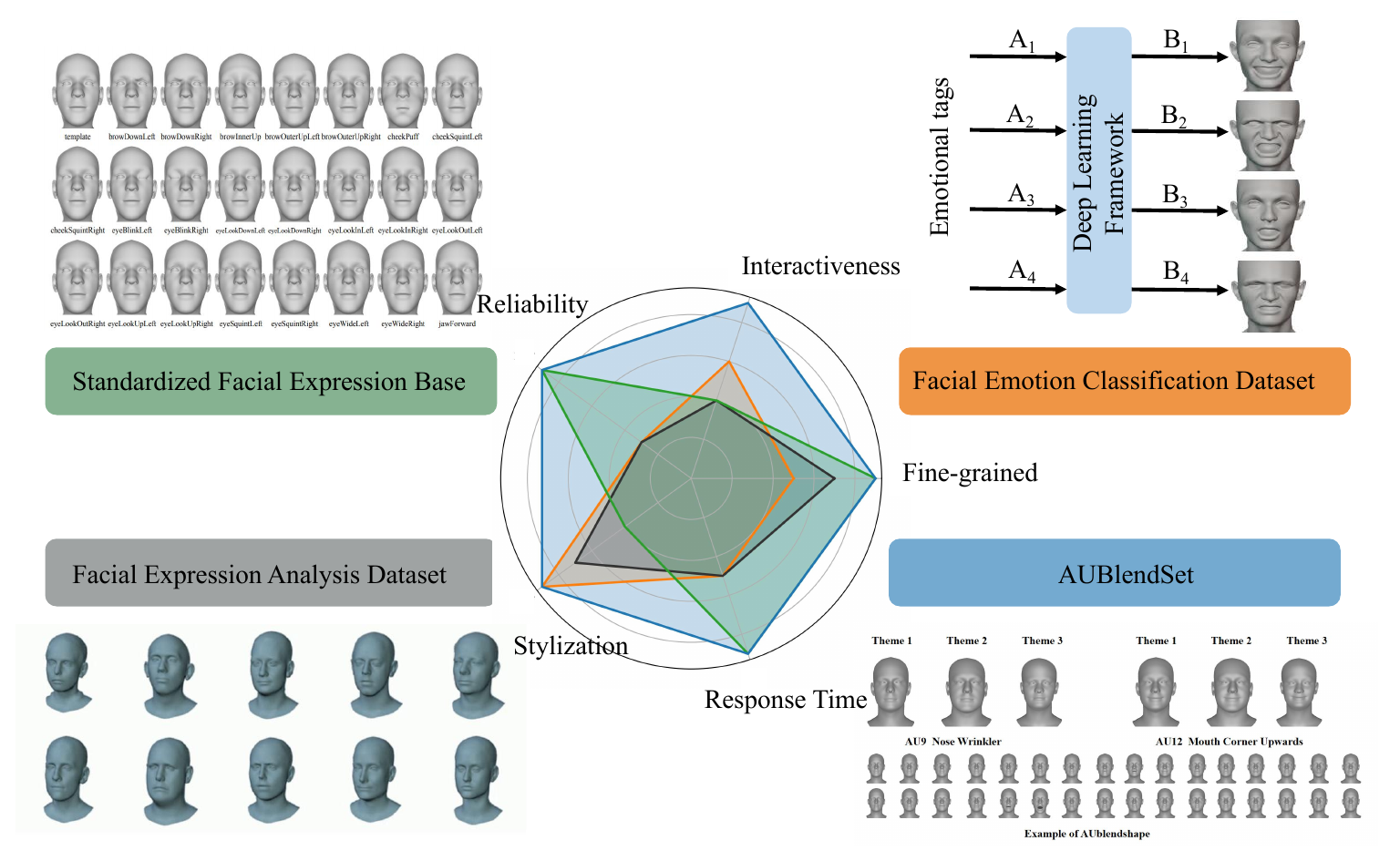}
 \vspace{-4mm}
 \caption{Comparisons of different datasets for facial expression manipulation regarding control basis reliability, stylization, interactiveness, fine-grained, and response time.}
 \label{introl}
\end{figure}

Due to diverse constraints in available facial emotion datasets, achieving fine-grained stylized facial expression manipulation confronts various challenges. As shown in Figure ~\ref{introl}, off-the-shelf standardized facial expression control usually generates characters' expressions through pre-defined universal blendshapes~\cite{KrumhuberTSR10,opneFACS}. Although those methods present some flexibility, they lack precise descriptions of stylization characteristics.

Facial emotion classification datasets~\cite{YinWSWR06,ZhangYCCRHLG14} typically annotate emotions into categories that do not provide appropriate facial control bases. Besides, the lack of subtle facial action annotations during the annotation process makes it challenging to achieve fine-grained facial expression manipulation.
Despite the facial expression analysis dataset using Facial Action Coding System (FACS)~\cite{ekman1978facial} to annotate facial details, it can not provide reliable facial manipulation like blendshape animation. Therefore, building a dataset with user-friendly interactiveness and support for stylized facial expression manipulation is essential and meaningful.

The FACS is widely adopted in behavioral science and psychology~\cite{ekman1978facial}. Based on facial muscle movements, AU-based representation offers a structured and interpretable framework capable of capturing the complexity of human expressions. 
with each AU representing an independent muscle. Facial expressions can be accurately described as combinations of corresponding AUs, and the FACS provides complete AU expression mapping rules. To support fine-grained and stylized 3D facial expression manipulation, we collect the facial expression control dataset, AUBlendSet, in AU-Blendshape format based on the expressive flame model. AUBlendSet contains 500 different identity themes. Each theme consists of 32 AU-Blendshape basis vectors of corresponding styles, accompanied by a set of facial poses with detailed AU annotations. Unlike existing public datasets, AUBlendSet assigns stylized AU-Blendshape control bases to each character for stylized facial expression manipulation.

Based on AUBlendSet, we propose the AUBlendNet for stylized facial expression manipulation. Our key insight is to generate stylized AU-Blendshape control bases, so AUBlendNet focuses on establishing the relationship between character identity and stylized AU-Blendshape control basis. Specifically, AUBlendNet first establishes a stylized AU-Blendshape control basis codebook, which uses the template mesh of the character as control conditions and predicts feature tokens using a non-autoregressive model. These tokens are leveraged to query the codebook sequence in discrete space to reconstruct the corresponding style control basis. Thanks to a large amount of training data from different identities, AUBlendNet is capable of quickly predicting the AU control basis for the given template mesh style and achieving fine-grained facial manipulation. We comprehensively validate the abilities of AUBlendSet and AUBlendNet through diversified downstream tasks, demonstrating their potential and value in 3D facial animation tasks. The main contributions are summarized as follows:

\begin{itemize}
    \item We collect the AUBlendSet dataset for stylized emotional facial animation. It includes stylized AU-Blendshape control bases and detailed facial AU poses to support stylized 3D facial expression manipulation.

    \item We propose the AUBlendNet model guided by character template meshes to concurrently predict all AU-Blendshpe control bases, which can perform fine-grained facial expression manipulation for any identity style.

    \item We formulate diverse testing tasks to demonstrate the importance and applicability of AUBlendSet dataset and the powerful capability of AUBlendNet model.
\end{itemize}

\section{Related Work}
\label{sec:related work}

\subsection{Facial manipulation with expressions}
Facial emotion manipulation has broad applications in virtual humans, movie animations, and emotion computing~\cite{GuoG22,AzariL24}. Extensive endeavors have been devoted to this task. ARKit~\cite{MenzelBL22} provides detailed descriptions of faces through 51 universal predefined blendshapes. However, with a large number of control bases and a lack of clear mapping rules for expression, it is problematic to achieve efficient, user-friendly facial manipulation. The FACS mapping table defines the mapping rules between AU and facial expression, which enables FACSGEN~\cite{KrumhuberTSR10} with AU as the control basis vector to modulate the corresponding expression quickly. However, due to the consistent style of the generated effects of the universal control basis, the obtained characters lack personalized characterizations. Li et al.~\cite{LiLY11} achieve personalized facial expression manipulation to some extent by assigning mesh weights to predefined controllers. However, this method is restricted by adaptability and has limited generalization ability when facing new facial topology structures. Some deep learning methods~\cite{PengWSXZH0F23,LG-LDM} use category labels as generation conditions to guide characters in generating corresponding facial emotions. Although characters can generate facial expressions of corresponding styles, the finite and discrete category labels significantly diminish the diversity and freedom of editing. Therefore, considering the current state of research, investigating facial expression manipulation techniques that have good interaction and can achieve fine-grained stylization is valuable and compelling.

\subsection{Facial datasets with expressions}
Existing facial datasets with expressions mainly focus on expression analysis and emotion calculation. AffectNet~\cite{MollahosseiniHM19} contains approximately 400,000 facial expression images by manually labeling eight expressions, as well as valence and arousal. CK+~\cite{ck} uses facial videos to record the expression changes from neutral to 7 basic expressions. This dataset comprises detailed facial keypoints and annotations of some AUs. DISFA~\cite{DISFA}  triggers spontaneous facial expression changes in participants and labels facial action units (AUs) with different intensities. FEAFA+~\cite{FEAFA,feafa+} provides a high-quality image collection with detailed annotations for 24 AUs. Despite 2D datasets being widely used due to their ease of access, annotations often lean on category labels, which cannot be directly mapped to independent expression control parameters, limiting their use in fine-grained facial expression manipulation.

The BU-3DFE \cite{YinWSWR06} dataset utilizes scanning devices to capture seven facial expressions of different subjects and annotate their intensity, which is widely used in facial expression modeling, emotion analysis, and 3D expression generation. However, this dataset has no precise annotation of facial details. BP4D~\cite{ZhangYCCRHLG14} and D3DFACS~\cite{D3DFACS} use FACS to describe facial detail information. However, each sample corresponds to multiple AU annotations, resulting in strong coupling relationships between different AUs. This makes it challenging to obtain stylized emotional facial manipulation. Although FaceWarehouse~\cite{FaceWarehouse} includes blendshape control bases for each character, its base vectors are entangled and can only be edited to a limited degree. Li et al.\cite{LiBZCIXRPKXL20} collect 4,000 high-resolution facial scans for character identity generation. However, this dataset utilizes a universal ARKit control base for emotional expression, with no style descriptions for character. To sum up, existing datasets have difficulty supporting fine-grained stylized facial expression manipulation in terms of data annotation, collection methods, and data formats.

\subsection{Applications of facial expression manipulation}
Facial emotion manipulation adjusts the emotional expression of digital characters through user editing. Shao et al.~\cite{ShaoZTLM21} propose a multi-class adversarial training method that achieves fine-grained expression transfer between unpaired images by decoupling AU-related and AU-unrelated features of facial images. Bodur et al.~\cite{BodurBK21} leverage depth information and depth consistency loss to manipulate facial expressions and combined confidence regularization to improve generalization ability. In speech-driven facial generation tasks, a large amount of work~\cite{PengWSXZH0F23,WangWSYWQHQL20,LG-LDM} uses one-hot encoding to generate facial expressions. However, due to the diversity of emotional expressions of different identities, the same emotion may correspond to different facial expression details. One-hot encoding cannot capture these fine-grained differences and often tends towards averaging during the training process, resulting in a lack of naturalness and personalization in the generated expressions. Meanwhile, in emotion and AU recognition tasks, automated annotation of AU can provide larger-scale, high-quality training data, thereby improving the accuracy and generalization ability of AU recognition and emotion analysis. We demonstrate that AuBlendSet can offer a rich collection of expression data to facilitate and enhance these applications.

\section{AUBlendSet}
\label{sec:formatting}
Blendshape is widely used in facial manipulation for its efficiency and controllability. To achieve fine-grained and stylized facial manipulation, we collect the AUBlendSet dataset, which encodes expressions using stylized AU-Blendshape representations. we first perform sampling in the FLAME parameter space based on key style factors (gender, age, and facial shape proportions), and systematically generate 500 characters to ensure comprehensive coverage of style variations. The dataset comprising 206 females and 294 males with age distributions ranging from 20 to 60 years. We adopt the expressive FLAME model to represent each character, where each mesh contains 5,023 3D vertices.

\begin{figure}[htbp]
 \centering 
 \includegraphics[width=1\columnwidth]{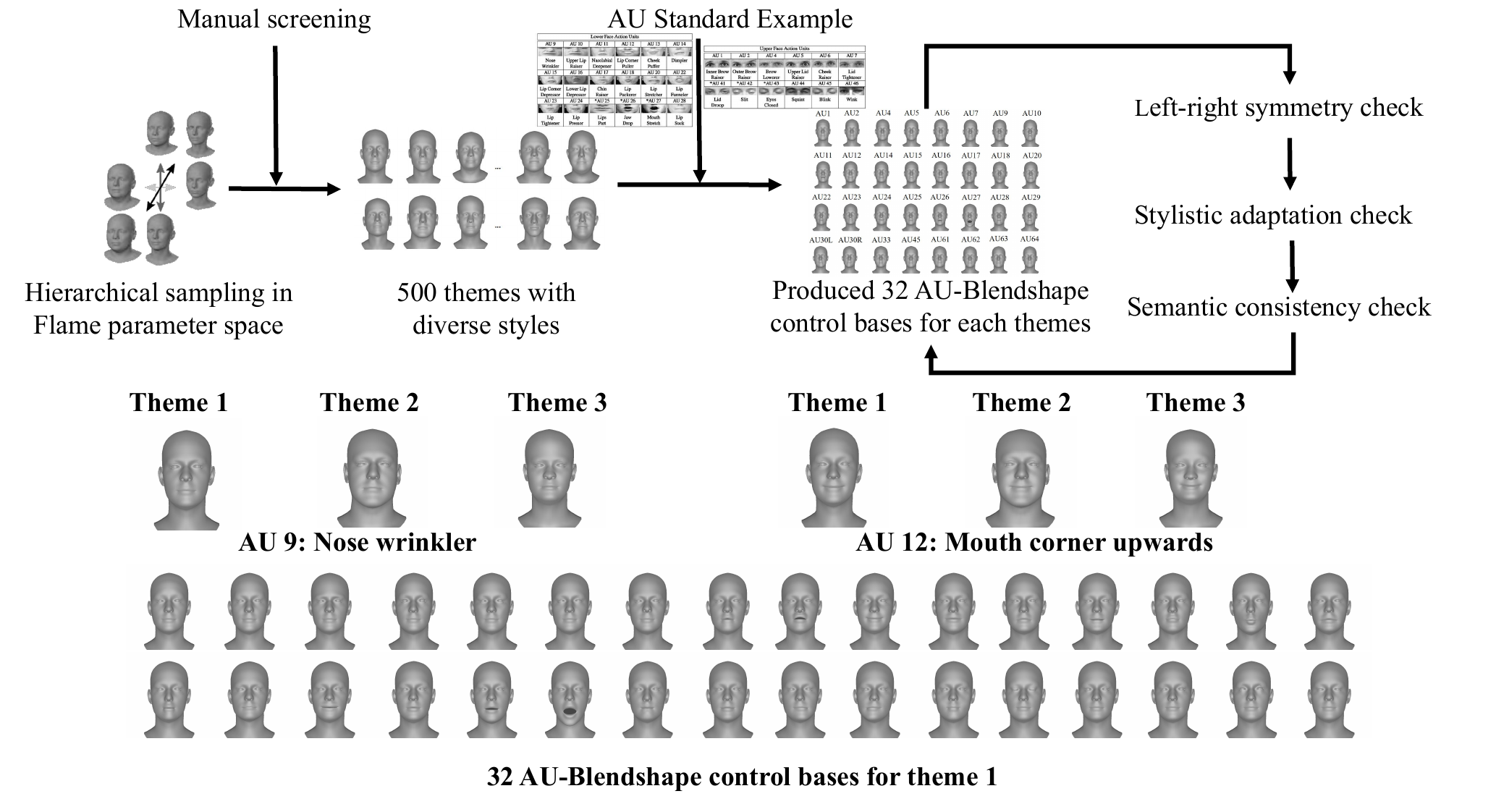}
  \vspace{-2mm}
 \caption{AUBlendSet production flow and examples. The top figure illustrates the production flow of the Blendshape. The bottom figure visualizes AU 9 (nose wrinkle) and AU 12 (mouth corner upwards) across three different themes, alongside the full set of 32 AU-Blendshape for Theme 1.}
 \label{aublendshape}
\end{figure}

The facial AU describes facial muscle movement and provides a reasonable interactive control basis for facial expression manipulation. Users can quickly manipulate and generate corresponding target expressions using the expression mapping rules defined in the FACS coding table~\cite{ekman1978facial}. However, due to the high similarity of some eye AU movements, they only involve varying degrees of movement. To achieve maximum fine-grained expression manipulation while ensuring good interaction, we simplify the description of eye AU to reduce redundancy while removing the AU descriptions of tongue and head posture. Thus, AUBlendSet ultimately retains 32 AU control bases. \textit{For specific AUs and FACS table, refer to the supplementary material (SM).} 


\begin{table}[htp]
\centering
\caption{Statistics of key indicators in the dataset.}
\vspace{-3mm}
\resizebox{0.48\textwidth}{!}{
\setlength\tabcolsep{2pt}
\begin{tabular}{c |c| c| c |c |c}
\hline
Name &Identifiers &Blendshape &AU Annotation &Annotation Type &Editing Suitability   \\
\hline 
FACSD3D~\cite{D3DFACS}  &10  &\ding{55} &\ding{51} &Discrete &low\\
CK+~\cite{ck}  &97 &\ding{55} &\ding{51} &Discrete &low\\
BP4D~\cite{BP4D}   &23  &\ding{55} &\ding{51} &Discrete &low\\
FaceWarehouse~\cite{FaceWarehouse} &150  &\ding{51} &\ding{55} &Continuous &low\\
AUBlendSet &500 &\ding{51} &\ding{51} &Continuous &high\\
\hline 
\end{tabular}
}
\label{keystatistics}
\end{table}

For each of the 500 characters in AUBlendSet, As shown in Figure \ref{aublendshape}, following standard animation production procedures, professional artists manually create blendshape control bases for each character, guided by AU exemplars defined in Facial Action Coding System (FACS). During this process, we ensure symmetry, stylistic adaptation, and semantic consistency with AU definitions,  with all results manually reviewed for clarity and structural correctness. This ensures that each theme contains a template mesh and 32 AU-Blendshape control bases that match the character's style. These control bases can accurately capture the unique facial dynamics of characters, making expression generation more natural and personalized. In addition, the dataset provides detailed annotated AU examples for each character theme, which can be used for expression generation and model performance evaluation, We present several key statistics of our dataset and related datasets in Table ~\ref{keystatistics}. supporting a wide range of experimental and analytical needs.


\section{AUBlendNet}
\begin{figure*}[htbp]
 \centering 
 \includegraphics[width=1.7\columnwidth]{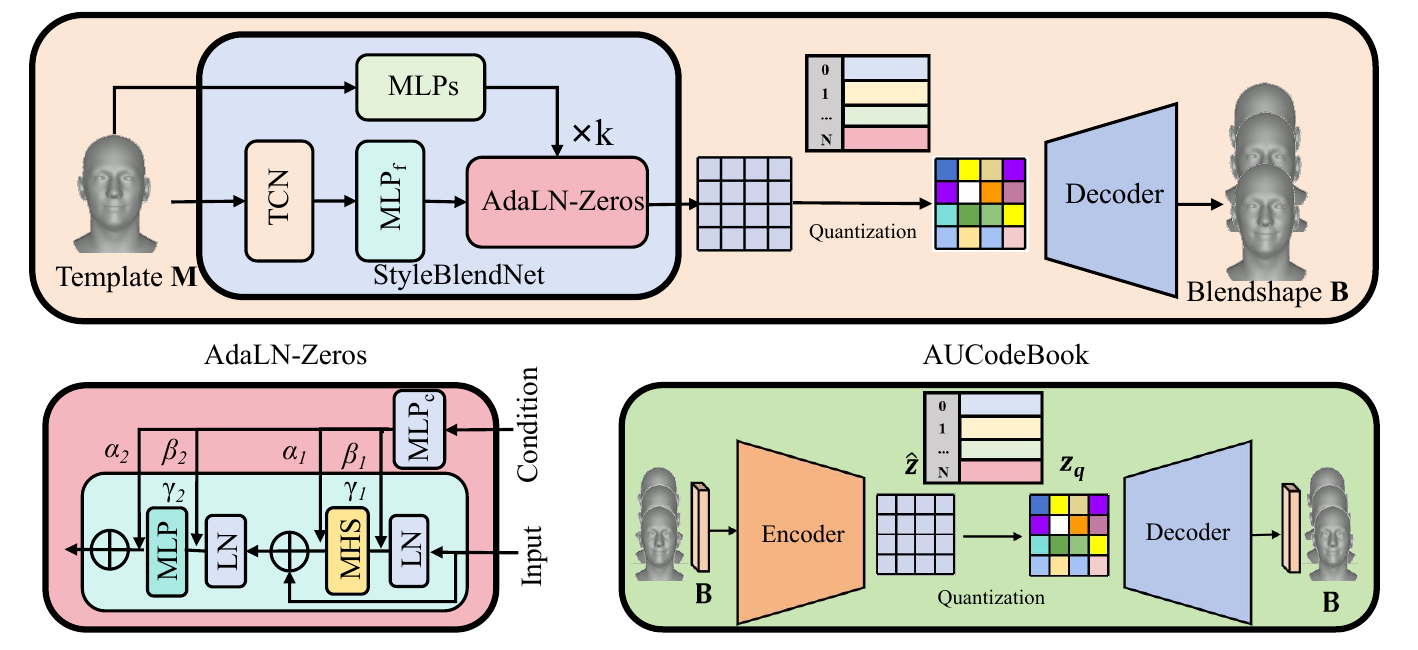}
  \vspace{-2mm}
 \caption{The pipeline of AUBlendNet, consisting of the AUCodebook and StyleBlendNet. AUCodebook is first pre-trained to construct an AU-style library in the hidden space. Given a character Template $\mathbf{M}$, StyleBlendNet first predicts stylized AU-Blendshape control bases code and then generates corresponding control bases $\mathbf{B}$ through the prior decoder in the AUCodeBook.}
 \label{pipline}
\end{figure*}

To achieve fine-grained, stylized, user-friendly interactive manipulation, the goal of AUBlendNet is to generate AU-Blendshape control bases that match the style of the character template, as shown in Figure \ref{pipline}. AUBlendNet mainly consists of the AUCodebook and StyleBlendNet. The AUCodebook constructs an AU-style library in the hidden space using learned motion prior information. The non-autoregressive StyleBlendNet maps the input facial template to the corresponding encoding vector in the Codebook in parallel, thereby predicting the AU-Blendshape control basis that matches the target character style.

\subsection{AUCodebook}
Inspired by CodeTalker\cite{codetalker}, our AUCodebook is a VAE-based Transformer architecture. It is optimized through a combined perceptual loss and a quantified training loss during training~\cite{OordVK17}. Precisely, given AU-Blendshape control bases $\mathbf{B} \in \mathbb{R}^{N \times V}$ for the identity in mesh space, the encoder $\mathcal{E}$ transforms AU-Blendshap base vectors $\mathbf{B}$ into latent representations $\mathbf{Z} = \mathcal{E}(\mathbf{B})$ and the decoder $\mathcal{D}$ reconstructs the AU-blendshape base vector from the latent representation $\mathbf{\hat{B}}  = \mathcal{D}(\mathbf{Z}) = \mathcal{D}( \mathcal{E}(\mathbf{B}$)), $\mathbf{Z} \in \mathbb{R}^{N \times D}$. $N$ is the number of AUs involved in the control, $V$ is the 5023 facial vertices, and $D$ is the latent space dimension.

\subsection{StyleBlendNet}
StyleBlendNet first expands the temporal dimension to fill in the missing time for input character templates. Subsequently, the extended sequence is projected onto the latent space, which can be described as:
\begin{equation}
\setlength\abovedisplayskip{3pt}
\setlength\belowdisplayskip{3pt}
\mathbf{\hat{Z}}=\rm{MLP_f}(\rm{TCN}(\mathbf{M})),
\end{equation}
where $\mathbf{M}\in \mathbb{R}^{1 \times W}$ is the mesh template for the input character, $\mathbf{\hat{Z}}\in \mathbb{R}^{N \times D}$ is the input token. $\rm{MLP}_f$ denotes the feature projection Module.

The style embedding module obtains the corresponding template style encoding, and we leverage the AdaLN-Zero block to inject the style encoding as a generation condition into the prediction process. The process is formulated as:
\begin{equation}
\setlength\abovedisplayskip{3pt}
\setlength\belowdisplayskip{3pt}
\boldsymbol{\gamma}_{1}, 
\boldsymbol{\beta}_{1}, \boldsymbol{\alpha}_{1}, \boldsymbol{\gamma}_{2}, \boldsymbol{\beta}_{2}, \boldsymbol{\alpha}_{2}= \rm{MLP}_{c}(\rm{{MLP}_s}(\mathbf{M})),
\end{equation}
where $\rm{MLP}_s$ denote style embeding, and $\rm{MLP}_c$ is the AdaLN-Zeros block. The control parameters $\boldsymbol{\gamma_{1}}$, $\boldsymbol{\beta_{1}}$, $\boldsymbol{\alpha_{1}}$, $\boldsymbol{\gamma_{2}}$, $\boldsymbol{\beta_{2}}$, $\boldsymbol{\alpha_{2}}$ are then integrated into a prediction network with a transformer backbone using AdaLN-Zero~\cite{DIT}:
\begin{equation}
\setlength\abovedisplayskip{3pt}
\setlength\belowdisplayskip{3pt}
\mathbf{\hat{Z}}_{t,k'} = \mathbf{\hat{Z}}_{t,k-1}+\boldsymbol{\alpha}_{1} \rm{MHSA}(\rm{LN}(\mathbf{\hat{Z}}_{t,k-1})\boldsymbol{\gamma}_{1}+\boldsymbol{\beta}_{1}),
\end{equation}
\begin{equation}
\setlength\abovedisplayskip{3pt}
\setlength\belowdisplayskip{3pt}
\mathbf{\hat{Z}}_{t,k} = \mathbf{\hat{Z}}_{t,k'}+\boldsymbol{\alpha}_{2}\rm{MLP}(\rm{LN}({\mathbf{\hat{Z}}_{t,k'}}) \boldsymbol{\gamma}_{2}+\boldsymbol{\beta}_{2}),
\end{equation}
where $\rm{MHSA}$ denotes multi-head self-attention, $\rm{LN}$ refers to layer normalization~\cite{VaswaniSPUJGKP17}. $\mathbf{\hat{Z}}_{t,k-1}$ and $\mathbf{\hat{Z}}_{t,k}$ are the outputs of $(k-1)$-th and $k$-th layer of prediction network. ${\mathbf{\hat{Z}}_{t,k'}}$ is the intermediate variable in the $k$-th layer of the network.

\subsection{Loss functions}
We first train the AUCodebook. Similar to CodeTalker~\cite{codetalker}, we adopt a facial mesh loss and two intermediate latent-level losses to optimize our AUCodeBook:
\begin{equation}
\setlength\abovedisplayskip{3pt}
\setlength\belowdisplayskip{3pt}
\mathcal{L}_{auc}\!=\!\|\mathbf{B}\!-\!\mathbf{\hat{B}}\|_1 \!+\!\|{\rm{SG}(\mathbf{\hat{Z}}})\!-\!\mathbf{Z}_{q}\|_2^2\!+\!\beta\|\mathbf{{\hat{Z}}}\!-\!\rm{SG}(\mathbf{Z}_{q})\|_2^2,
\end{equation}
where the first term is a facial mesh reconstruction loss, $\mathbf{\hat{Z}}$ is the AU-Blendshape embedded features. $\mathbf{Z}_{q}$ is the quantized AU-Blendshape embedded features, $\rm{SG}$ stands for a stop-gradient operation, and $\beta$ refers to the weighting factor controlling the update rate.

Secondly, we start to optimize the StyleBlendNet. We leverage a facial mesh loss and an intermediate latent-level loss to train StyleBlendNet:
\begin{equation}
\setlength\abovedisplayskip{3pt}
\setlength\belowdisplayskip{3pt}
\mathcal{L}_{syn}=\|\mathbf{B}-\mathbf{{\hat{B}}}\|_2^2+\|\mathbf{\hat{Z}}-\rm{SG}(\mathbf{Z}_{q})\|_2^2.
\end{equation}

\section{Experiments}

We explore the capabilities and potential of AUBlendSet and AUBlendNet through diverse tasks, including stylized facial expression manipulation, speech-driven emotional facial animation, and AUBlendNet for AU detection.


\subsection{Implementation details}
 Our framework is built on the PyTorch platform and trained on RTX 6000 Ada. First, we optimize the AUCodebook, setting its layer numbers to 8 and latent spatial dimensions to 1024. The hyperparameter $\beta$ equals 0.1. Adam optimizer is used to train AUCodebook with the learning rate  1$\times$ $10^{-4}$, and a batch size of 1 for 200 epochs. Subsequently, we optimize StyleBlendNet by setting its hidden space dimension to 1024. The head number and layer number are both set to 8. We train StyleBlendNet with a batch size of 1 for 400 epochs and a learning rate set of 1 $\times$ $10^{-5}$.

\subsection{Stylized facial expression manipulation}
Although the universal facial blendshape control base can be applied to most tasks, in more realistic character expression manipulation processes, it often fails to reflect the unique style of character emotions. The key to stylized facial expression manipulation lies in constructing control bases that can adapt to each character's personalized characteristics and emotional expressions, thereby achieving natural and accurate emotional style shaping for any character.

\textbf{Experiments on different datasets for expression manipulation.}
 To validate the superiority of AUBlendSet with AU-Blendshape representation for stylized facial expression control, we implement AUBledNet on different datasets to validate the generation effects. We select two representative datasets with AU annotation with our dataset for comparison. 
The D3DFACS~\cite{D3DFACS} is constructed based on the FLAME model. It consists of 110 single-AU and 352 multi-AU combination data for 10 characters.
 Each piece of data gradually changes from a neutral posture to a fully activated state to accurately capture the dynamic changes in facial expressions. The CK+~\cite{ck} contains 97 different characters, each with at least 5 different expressions and accompanied by AU annotations. CK+ has 55 single-AU and 435 multi-AU combination annotations, providing rich annotation data for facial expression analysis. We leverage EMOCA~\cite{emoca} to transform 2D images into 3D Flame model for subsequent model training. 
 Since D3DFACS~\cite{D3DFACS} and CK+ contain AU labels, we inject AU labels as a generation condition into the prediction process for fair comparisons. We divide each dataset into training, validation, and testing sets with an 8:1:1 ratio and train them using AUBlendNet.

 \begin{table}[htp]
\centering
\caption{Comparison results of AUBlendNet trained on D3DFACS, CK+, and AUBlendSet on the AUBlendSet test set.}
\begin{tabular}{c | c c c }
\hline
\multirow{2}{*}{Method}     &MSE$_{S}$\(\downarrow\) &MES$_{M}$\(\downarrow\) &Inference \(\downarrow\) \\
                            &($\times 10^{-9}$)&($\times 10^{-9}$) &(s)\\
\hline 

D3DFACS\cite{D3DFACS}  &226.41	&68.21	 &2.7\\
CK+\cite{ck} &8490.27 	&2028.14   &2.7\\
AUBlendSet  &\textbf{4.55}	&\textbf{7.62} &\textbf{0.3}\\
\hline
\end{tabular}
\label{dataset comparison}
\end{table}
 \begin{figure}[htb]
 \centering 
 \includegraphics[width=\columnwidth]{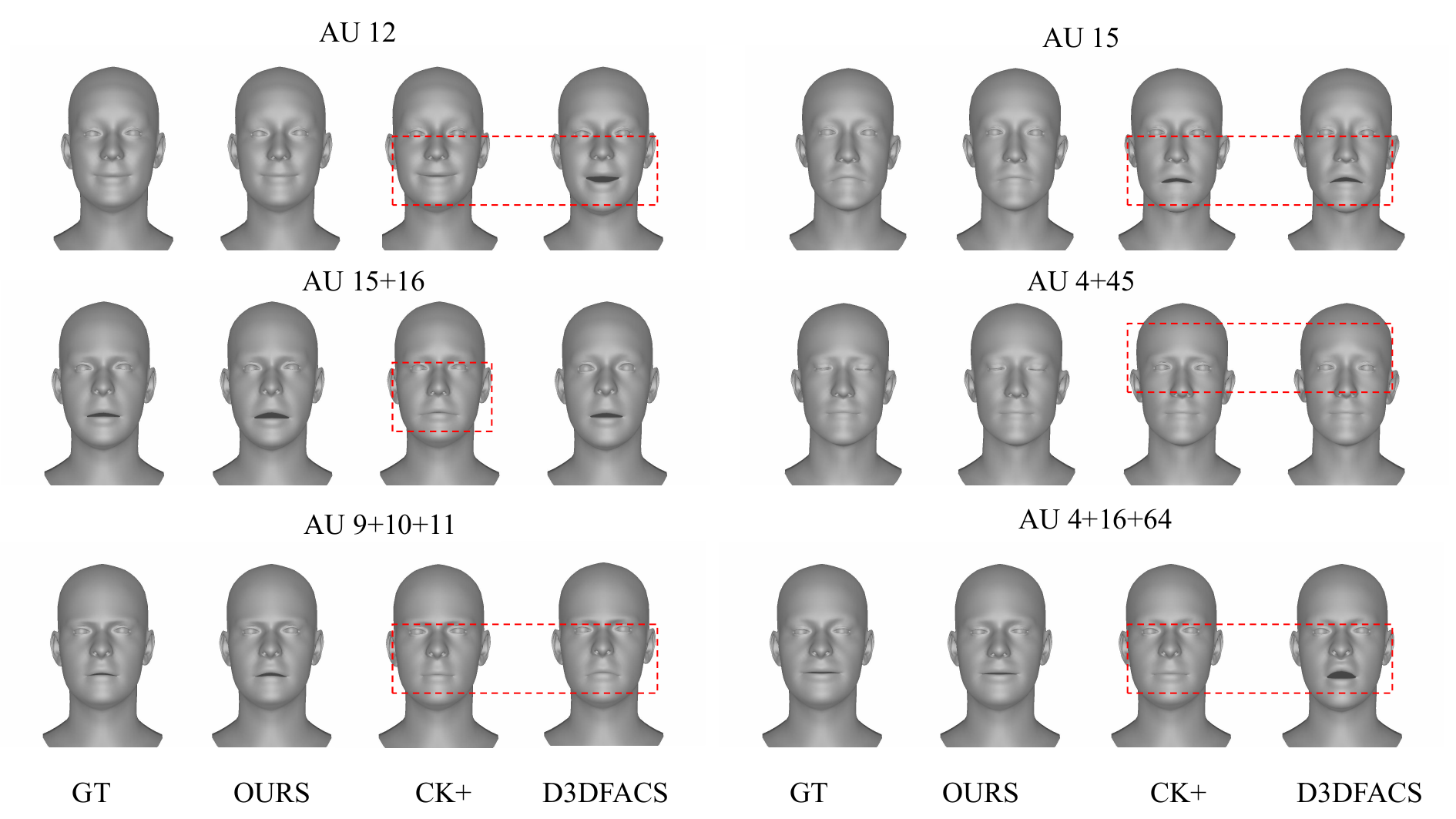}
 \vspace{-4mm}
 \caption{ Visualization Comparisons of AU-guided emotional face generation results for AUBlendNet trained on different datasets.}
 \label{dataset-comp}
\end{figure}

 Table~\ref{dataset comparison} records the results of AUBlendNet optimized by three datasets on the AUBlendSet test set. 
 MSE$_{S}$ denotes the generation error of a single-AU control, and MSE$_{M}$ is the generation error of multi-AU controls. 
 From Table \ref{dataset comparison}, we observe that although CK+  has more identities, D3DFACS contains more samples for each theme, resulting in higher control accuracy for D3DFACS than CK+. Due to the direct AU annotations in CK+ and D3DFACS, such enumeration labels make it difficult to avoid errors when facing multi-AU-driven control processes. Furthermore, we can find that our AUBlendSet dataset is much more effective in generating results for single-AU control and multi-AU collaborative control. The superior performance can be attributed to AU-Blendshape representation, which simplifies the process of generating controllable facial postures. In the face of complex multiple AUs control, we only need to blend the offsets generated by each AU-Blendshape. By comparing the CK and D3DFACS datasets, which rely on direct AU annotations, we observe that when encountering AU control signals not seen in the training set, the optimized AUBlendNet tends to lead to biased results, favoring those generations within the training set. This phenomenon can be observed intuitively in Figure \ref{dataset-comp}. The facial meshes generated from our AUBlendSet are much more consistent with the given control signals, whereas the CK+ and D3DFACS exhibit significant errors and unwanted motions. This continuous control enables fine-grained expression adjustments and smooth transitions, overcoming the limitations of discrete or static datasets and supporting more natural and flexible facial animation. Those quantitative and qualitative comparison results demonstrate the superiority and generalization of our AUBlendSet dataset.

\begin{figure}[htb]
 \centering 
 \includegraphics[width=\columnwidth]{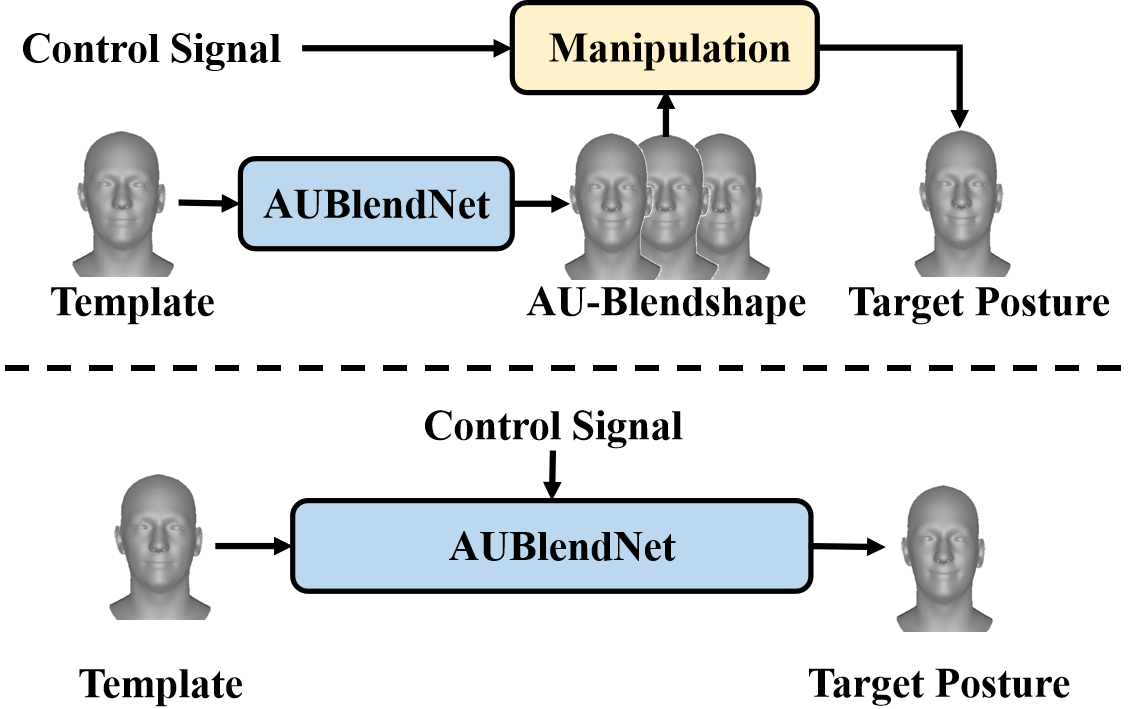}
 \vspace{-2mm}
 \caption{Workflow comparisons of AUBlendNet for AU-guided facial expression control. The up first predicts AU-Blendshape basis, then blends them with control signal to obtain target posture. The bottom directly predicts target posture given control signal. }
 \label{runtime}
\end{figure}

\textbf{Computation efficiency analysis.}
Inference generation speed is a crucial criterion for facial expression manipulation. We compare the workflows based on AU-Blendshape control basis and AU category labels in Figure ~\ref{runtime}. The response times of facial postures for inferring 7 different emotions are reported in the last column of Table \ref{dataset comparison}. For the CK+\cite{ck} and D3DFACS\cite{D3DFACS} datasets labeled with AU categories, given changed AU control signals, model requires individual predictions each time, resulting in cumbersome and inefficient inference processes. In contrast, AUBlendNet trained on AUBlendSet with AU-Blendshape representations only requires one-step prediction to generate the corresponding style of AU-BlendShape base vectors and then conducts expression manipulation through blendshape different AU control signals. Specifically, on the RTX 6000 Ada, AUBlendNet trained on AUBlendSet takes 0.3 seconds to infer the character's stylized AU-BlendShape, while manipulation through AU-Blendshape basis takes only 0.002 seconds. However, the AUBlendNet trained with category labels on CK+\cite{ck} and D3DFACS\cite{D3DFACS} takes 2.7 seconds to infer 7 emotional postures. This indicates that AUBlendNet significantly improves the real-time efficiency of facial manipulation tasks by generating control bases and completing manipulation through blending AU-Blendshape bases.

\begin{table}[htbp]
\centering
\caption{Quantitative comparison of different models for AU-guided 3D facial expression manipulation on AUBlendSet.}
\setlength\tabcolsep{4pt}
\vspace{-2mm}
{
\begin{tabular}{c | c c c}
\hline
\multirow{2}{*}{Method}     &MSE$_{S}\downarrow$ &MES$_{M}\downarrow $ \\
                            &($\times 10^{-9}$)&($\times 10^{-9}$)\\ 
\hline 
Diffusion\cite{FaceDiffuser}	&5.66	&8.71\\	
VQ-VAE\cite{codetalker}	&8.48 	&12.79\\	
Transformer\cite{tansf}	&86.12	&96.14\\	
AUBlendNet	&\textbf{4.55}	&\textbf{7.62}\\ 
\hline
\end{tabular} }
\label{comparison}
\end{table}

\textbf{Comparisons with other models.}
We validate the superiority of AUBlendNet for stylized facial expression control. As far as we know, there are no available methods for AU-guided 3D facial expression manipulation in deep learning. Thus, we implement the current state-of-the-art conditional generation models based on our collected AUBlendSet. The contrast models consist of a Transformer~\cite{VaswaniSPUJGKP17}, a VQ-VAE~\cite{codetalker}, and a Diffusion model~\cite{FaceDiffuser}. All compared methods establish the relationship between identity mesh and the AU-Blendshape basis vectors conditioned on character styles. We also utilize MSE$_{S} $ and MSE$_{M}$ as evaluation metrics to quantitatively analyze the generation quality of single-AU and multi-AU control signals. Table~\ref{comparison} illustrates the comparison results. From Table~\ref{comparison}, we observe that the proposed AUBlendNet significantly outperforms the compared methods in both single-AU and multi-AU facial expression manipulation.

\begin{figure}[htb]
 \centering 
 \includegraphics[width=0.9\columnwidth]{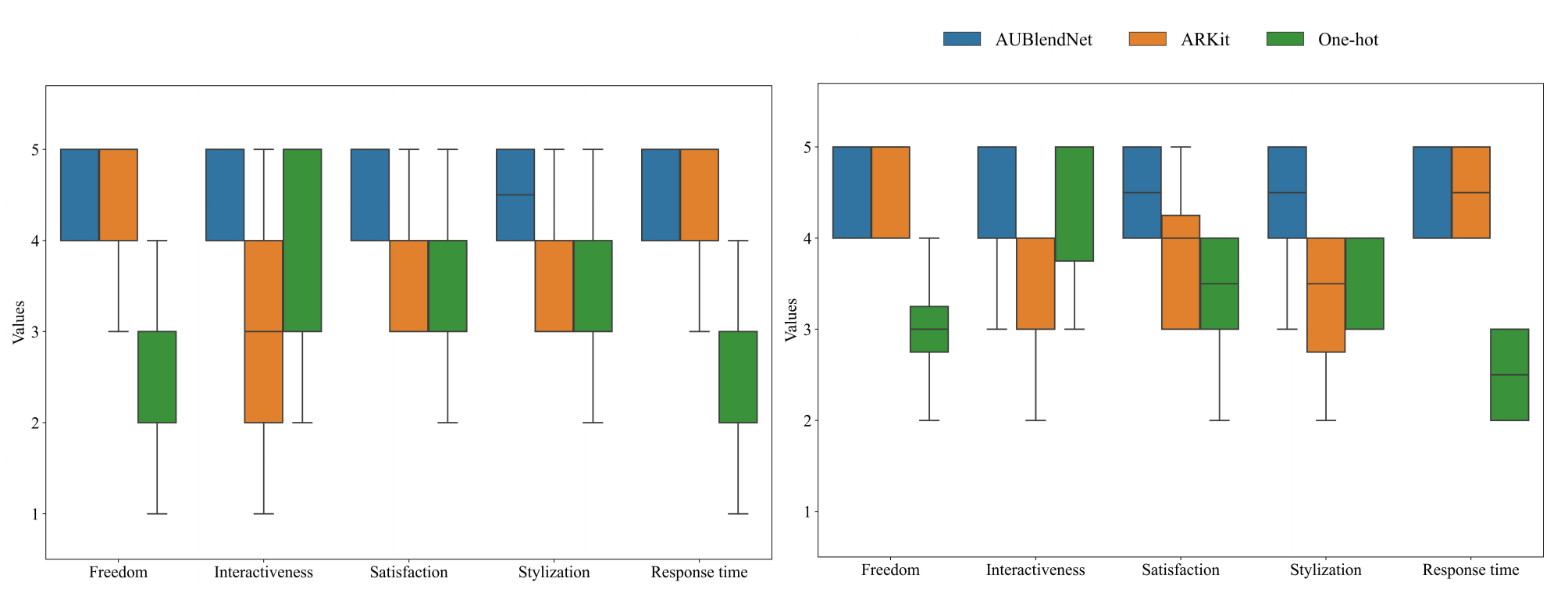}
 \vspace{-2mm}
 \caption{User study evaluation for facial expression manipulation. Right is 8 animation experts, Left is 32 general participants.}
 \label{userstudy}
\end{figure}

\textbf{User study evaluation.}
Controllability and generative validity are crucial for evaluating model performance. To compare different methods, we recruited 32 general participants and 8 animation experts to evaluate the emotional face generation results of the ARKit-based method~\cite{MenzelBL22}, the one-hot-based method~\cite{LG-LDM}, and our method. Participants were asked to edit facial expressions (e.g., happiness, anger) using different methods and provide qualitative ratings. We use a 5-point scale to assess different methods regarding manipulation freedom, interactiveness, generation satisfaction, style consistency, and response time (\textit{Refer to SM}). We report the statistical results in Figure \ref{userstudy}. 

\begin{figure*}[ht]
 \centering 
 \includegraphics[width=1.75\columnwidth]{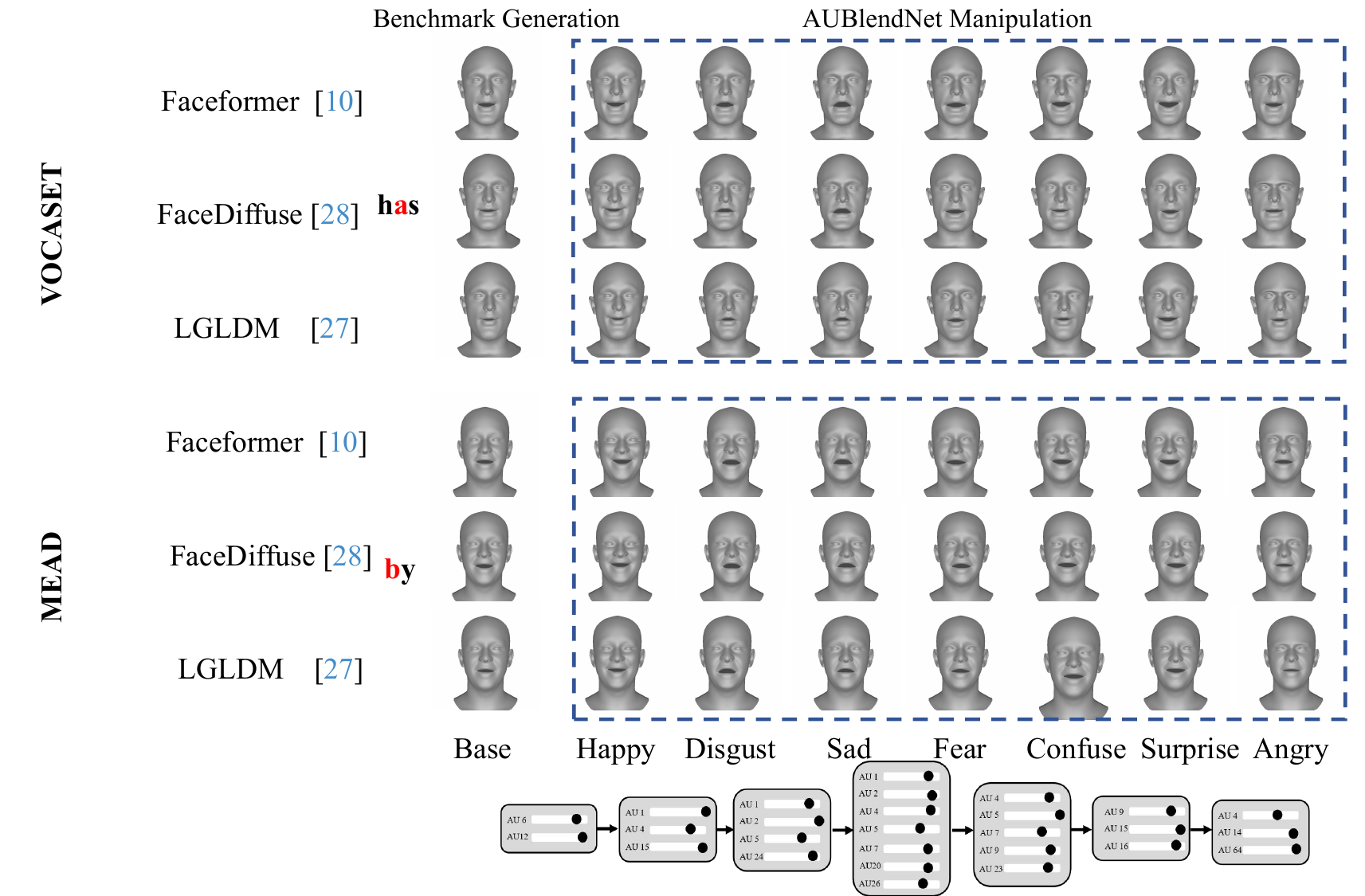}
 \vspace{-2mm}
 \caption{Visualization results of different speech-driven methods combined with AUBlendNet. The red font represents the speech information corresponding to the current frame. The lower section displays fine-grained AU-driven facial expression manipulation parameters.}
 \label{emo_edit}
\end{figure*}

The experimental results indicate that the recruited participants and animation experts prefer our AUBlendNet, which performs well for all the criteria. 
Users can quickly manipulate corresponding emotions through the FACS coding table, and the offline generation of stylized control bases significantly enhances the user's interactive experience. Thanks to AUBlendNet's direct prediction of AU-BlendShape that matches the character's style, our method is comparable to the ARKit-based BlendShape method in terms of real-time and interactivity. At the same time, the AU-BlendShape control bases generated by AUBlendNet not only accurately match the corresponding AU but also conform to the style characteristics of the corresponding character, ensuring the naturalness and consistency of the generated results.

\subsection{Speech-driven emotional facial animation}
In existing emotional speech-driven facial animation methods, expressions are usually injected as category labels into the generation process, often leading to almost consistent expression results among all characters. However, characters often exhibit diversity and personalization in their emotional expressions. A single expression category label makes it difficult to generate unique emotional responses for different characters. To address this issue, we demonstrate an intriguing speech-driven emotional facial motion generation method based on AUBlendSet and AUBlendNet.


Through observation, we find that the leading facial changes are concentrated on the lips, and the expression of character emotions mainly focuses on the upper face and corners of the mouth in a speech-driven process. This enables the AUBlendNet predicted AU-Blendshape control basis to be easily integrated into common speech-driven facial frameworks for facial expression manipulation through blendshape control basis. This process can be described as:
\begin{equation}
\setlength\abovedisplayskip{1pt}
\setlength\belowdisplayskip{1pt}
\mathbf{M}=\mathbf{M_0}+\mathbf{Offset_{spe}}+\mathbf{Offset_{Exp}},
\end{equation}
where $\mathbf{M}$ is the final facial posture, $\mathbf{M_0}$ is the facial template posture, $\mathbf{Offset_{spe}}$ is the lip deviation driven by speech, and $\mathbf{Offset_{Exp}}$ is the facial expression deviations based on AU-Blendshape representation.

\textbf{Visualization analysis.}
We select three representative speech-driven models, Faceformer~\cite{fan2022faceformer}, FaceDiffuser~\cite{FaceDiffuser}, and LGLDM~\cite{LG-LDM} as benchmarks and compare the emotional generation results with and without our AUBlendNet. As shown in Figure~\ref{emo_edit}, we demonstrate the neutral expression (Base) generation results without AUBlendNet. Meanwhile, we illustrate the fine-grained continuous manipulation effects of the AU-Blendshape control bases predicted by AUBlendNet on seven emotions. The stylized AU-Blendshape control base supports the adjustment of the corresponding AU and its activation intensity, making facial expression control more refined. For VOCASET (without emotion labels), AUBlendNet achieves continuous manipulation of emotions based on the FACS coding table, generating diverse facial expressions. For MEAD (with emotion labels), such as happy expressions, compared to the generated results of using One-hot labels, the programmatic control base provided by AUBlendNet can not only finely adjust the current emotion but also flexibly achieve natural transitions between different emotions.

\begin{table}[t]
\centering
\caption{Comparisons of different speech-driven models with and without combined with AUBlendNet on MEAD and VOCASET. ``N/A'' denotes Not Applicable, since VOCASET has no emotions.}
\setlength\tabcolsep{4pt}
\vspace{-2mm}
\resizebox{1\linewidth}{!}
{
\begin{tabular}{c | c c c c }
\hline
\multicolumn{5}{c}{MEAD}\\
\hline
\multirow{2}{*}{Methods}     &LVE $\downarrow$ &FDD $\downarrow$ &V-LVE$\downarrow$ &Diversity $\uparrow$\\
                            &($\times 10^{-4}$)&($\times 10^{-6}$) &($\times 10^{-5}$) &($\times 10^{-6}$)\\
\hline 
Faceformer\cite{fan2022faceformer}	&\textbf{2.83}	&1.57 &2.64 &2.16\\
Faceformer+AUBlendNet	&3.04	&\textbf{1.55} &2.64  &\textbf{2.74}  \\
\hline
FaceDiffuse\cite{FaceDiffuser}	&\textbf{1.47}	&1.84 &3.11 &2.19 \\
FaceDiffuse+AUBlendNet	&1.76	&\textbf{1.81}  &3.11& \textbf{2.67}\\
\hline
LGLDM\cite{LG-LDM}	&\textbf{2.01} 	&\textbf{5.79}	 &2.15 &2.42 \\
LGLDM+AUBlendNet	&2.35 &5.96  &2.15 &\textbf{2.91} \\
\hline
\multicolumn{5}{c}{VOCASET}\\
\hline
\multirow{2}{*}{Methods}     &LVE $\downarrow$ &FDD $\downarrow$ &V-LVE$\downarrow$ &Diversity $\uparrow$\\
                            &($\times 10^{-4}$)&($\times 10^{-6}$) &($\times 10^{-5}$) &($\times 10^{-6}$)\\
\hline 
Faceformer\cite{fan2022faceformer}	&\textbf{4.11}	&\textbf{4.67}  &2.91 &N/A\\
Faceformer+AUBlendNet	&4.45	&4.96  &2.91 &\textbf{2.17}\\
\hline
FaceDiffuse\cite{FaceDiffuser}	&\textbf{3.79}	&4.26 &4.08 &N/A\\
FaceDiffuse+AUBlendNet	&4.04	&\textbf{4.22}  &4.08 &\textbf{2.06}\\
\hline
LGLDM\cite{LG-LDM}	&\textbf{3.79} 	&\textbf{3.92}	&3.12 &N/A \\
LGLDM+AUBlendNet	&3.93	&3.95  &3.12 &\textbf{2.31}\\
\hline 
\end{tabular}
 }
\label{speech-drive}
\end{table}

\textbf{Quantitative comparison.}
The stylized AU-Blendsh-ape control bases provided by AUBlendNet can achieve diverse facial expression manipulation. Following \cite{codetalker}, we adopt the commonly used evaluation metrics LVE (Lip Vertex Error), V-LVE (Velocity of Lip Vertex Error), and FDD (Upper-face Dynamics Deviation) to quantitatively evaluate the generated results. Meanwhile, we use the indicator proposed by Ren et al.\cite{RenPZK23} to evaluate the diversity of facial expression generation results.

We evaluate the speech-driven facial emotional generation results based on 7 different emotions provided by FACS on VOCASET and MEAD. As shown in Table \ref{speech-drive}. The results indicate that due to the involvement of mouth corner movements in some AUs, the LVE performance decreases slightly during facial emotion manipulation, but this does not affect the offset of lip movements velocity. In the V-LVE and FDD indicators, it can be seen that using AU-Blendshape to manipulate facial emotions can ensure lip synchronization consistency and preserve the character's speaking style. However, the additional facial emotion manipulation significantly enhances the diversity and emotional expressiveness of facial expressions, resulting in more prosperous and natural emotional responses, further enhancing the realism and interactivity of facial animations.

\subsection{AUBlendNet for AU detection}
In facial AU detection, manually annotating activated facial AUs is time-consuming and costly, restricting dataset expansion and diversity. Capturing high-quality facial geometric information requires professional equipment, while fine-grained AU annotation demands considerable human effort. 
To address the issue, we leverage AUBlendNet as a data augmentation tool. AUBlendNet is capable of effectively improving AU detection performance while enhancing the model's generalization ability and robustness.

\begin{table}[t]
\centering
\caption{Comparisons of different AU detection methods with and without AUBlendNet-based data augmentation.}
\setlength\tabcolsep{4pt}
\vspace{-2mm}
\begin{tabular}{c | c c c c }
\hline

\multirow{2}{*}{Methods}     &D3DFACS &DISFA \\
                            &F1 score $\uparrow$ &F1 score $\uparrow$ \\
\hline 
HMP-PS\cite{SongCZJ21}	&55.04	&61.96 \\
HMP-PS + AUBlendNet	&\textbf{57.96}	&\textbf{64.57} \\
\hline
LP-Net\cite{NiuHYHS19}	&52.76	&56.61	\\
LP-Net + AUBlendNet	&\textbf{54.96}	&\textbf{59.35} \\
\hline
SRERL\cite{LiZZWL19}	&52.48 	&55.75	\\
SRERL + AUBlendNet	&\textbf{55.96}	&\textbf{56.32} \\
\hline
\end{tabular}
\label{AU-detection}
\end{table}

We demonstrate the effectiveness of AUBlendNet as a data augmentation tool for AU recognition. We compare the experimental results of HMP-PS~\cite{SongCZJ21}, LP-Net~\cite{NiuHYHS19}, and SRERL~\cite{LiZZWL19} on D3DFACS and DISFA~\cite{DISFA} datasets. For the DISFA dataset, we convert it to the Flame model using EMOCA~\cite{emoca} and train it according to the experimental setup in the paper. The accuracy of AU recognition is evaluated by the average F1 score of 8 AUs (AU1, AU2, AU4, AU6, AU9, AU12, AU25, AU26). Results are shown in Table~\ref{AU-detection}. By generating stylized AU-Blendshape, we construct more diverse AU combinations based on the original data, thereby enhancing the generalization ability of AU recognition algorithms to different character styles and expression changes. The experimental results show that using AUBlendNet-based augmented data to train the AU recognition model improves recognition precision on different datasets, especially in the case of a small amount of annotated data. The augmented data effectively alleviates the problem of high AU annotation cost, further verifying the practicality of AUBlendNet in 3D facial animation tasks.

\section{Conclusion}
In this paper, we contribute a facial action unit blendshape guided dataset, AUBlendSet, for fine-grained stylized facial expression manipulation. AUBlendSet has 500 identity themes, and each contains 32 AU-Blendshape basis vectors and corresponding facial posture sets with detailed AU annotations. Meanwhile, we propose AUBlendNet for generating stylized AU-Blendshape control bases. Given the identity mesh, AUBlendNet predicts the corresponding style of AU-Blendshape basis vectors. We achieve fine-grained stylized acial expression manipulation by linearly blending AU-Blendshape control bases. We thoroughly evaluate the capabilities of AUBlendSet and AUBlendNet through diverse tasks, including stylized facial expression manipulation, speech-driven emotional facial animation, and leveraging AUBlendNet for AU detection. Extensive experiments demonstrate the potential and importance of AUBlendSet and AUBlendNet in 3D facial animation tasks. 

\section*{Acknowledgement}
This research is supported in part by National Key R${\&}$D Program of China (No. 2022ZD0115902), the Major Key Project of PengCheng Laboratory (PCL2023A10-2), Beijing Natural Science Foundation(4232023), China Ministry of Education Funds for Humanity, Social Sciences (24YJCZH458) and the Pioneer Centre for AI, DNRF grant number P1.

{
    \small
    \bibliographystyle{ieeenat_fullname}
    \bibliography{main}
}

\end{document}